\documentclass{article}
\PassOptionsToPackage{hidelinks}{hyperref}
\PassOptionsToPackage{hyphens}{url}

\usepackage{arxiv}
\usepackage{enumitem}
\usepackage[utf8]{inputenc}
\usepackage[T1]{fontenc}
\usepackage{booktabs}
\usepackage{amsmath,amssymb,amsfonts}
\usepackage{nicefrac}
\usepackage{microtype}
\usepackage{lipsum}
\usepackage{algorithmic}
\usepackage{graphicx}
\usepackage{textcomp}
\usepackage{tabularx}
\usepackage{xcolor}
\usepackage{listings}
\usepackage[numbers,sort&compress]{natbib}
\usepackage{doi}
\usepackage{fancyhdr}
\usepackage{subcaption}
\usepackage{stfloats}
\usepackage{pifont}
\usepackage{cleveref}
\newcommand{\xmark}{\ding{55}}

\title{Avionic Main Fuel Pump Simulation and Fault-Diagnosis Benchmark}

\date{}

\newif\ifuniqueAffiliation
\uniqueAffiliationtrue

\ifuniqueAffiliation 

\author{
\mbox{Felix L. Janzen\hspace{0.3em}%
\href{https://orcid.org/0009-0000-5235-1767}{\raisebox{0.2ex}{\includegraphics[scale=0.06]{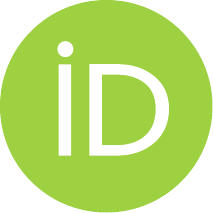}}}}\\
Institute of Artificial Intelligence\\
Helmut Schmidt University\\
Hamburg, Germany \\
\texttt{Felix.Janzen@hsu.hamburg}
\And
\mbox{Lukas Moddemann\hspace{0.3em}%
\href{https://orcid.org/0000-0002-9470-0811}{\raisebox{0.2ex}{\includegraphics[scale=0.06]{orcid.pdf}}}}\\
Institute of Artificial Intelligence\\
Helmut Schmidt University\\
Hamburg, Germany \\
\texttt{Lukas.Moddemann@hsu.hamburg}
\AND
\mbox{Alexander Diedrich\hspace{0.3em}%
\href{https://orcid.org/0000-0002-8674-6895}{\raisebox{0.2ex}{\includegraphics[scale=0.06]{orcid.pdf}}}}\\
Institute of Artificial Intelligence\\
Helmut Schmidt University\\
Hamburg, Germany \\
\texttt{Alexander.Diedrich@hsu.hamburg}
\And
\mbox{Oliver Niggemann\hspace{0.3em}%
\href{https://orcid.org/0000-0001-8747-3596}{\raisebox{0.2ex}{\includegraphics[scale=0.06]{orcid.pdf}}}}\\
Institute of Artificial Intelligence\\
Helmut Schmidt University\\
Hamburg, Germany \\
\texttt{Oliver.Niggemann@hsu.hamburg}
}
\else
\usepackage{authblk}

\newbox{\orcid}\sbox{\orcid}{\includegraphics[scale=0.06]{orcid.pdf}} 
\author[1]{%
	\href{https://orcid.org/0009-0000-5235-1767}{\usebox{\orcid}\hspace{1mm}David S.~Hippocampus\thanks{\texttt{hippo@cs.cranberry-lemon.edu}}}%
}
\author[1,2]{%
	\href{https://orcid.org/0000-0000-0000-0000}{\usebox{\orcid}\hspace{1mm}Elias D.~Striatum\thanks{\texttt{stariate@ee.mount-sheikh.edu}}}%
}
\affil[1]{Department of Computer Science, Cranberry-Lemon University, Pittsburgh, PA 15213}
\affil[2]{Department of Electrical Engineering, Mount-Sheikh University, Santa Narimana, Levand}
\fi

\renewcommand{\headeright}{}
\renewcommand{\undertitle}{}
\renewcommand{\shorttitle}{}

\hypersetup{
pdftitle={A template for the arxiv style},
pdfsubject={q-bio.NC, q-bio.QM},
pdfkeywords={First keyword, Second keyword, More},
}

\makeatletter
\newcommand{\ifacnotice}[1]{%
  \renewcommand{\footnoterule}{}
  \long\def\@makefntext##1{\noindent ##1}
  \footnotetext{%
    \noindent\fbox{%
      \parbox{\dimexpr\columnwidth-2\fboxsep-2\fboxrule\relax}{%
        #1
      }%
    }%
  }%
  \addtocounter{footnote}{-1}%
}
\makeatother

\begin{document}
\maketitle
\ifacnotice{%
  \vspace{1mm}
  © 2026 the authors. This work has been accepted to IFAC for publication under a Creative Commons Licence CC-BY-NC-ND
  \vspace{1mm} }

\thispagestyle{empty}

\begin{abstract}
	In many cyber-physical systems, especially in critical applications such as aeroplanes, data to train anomaly detection and diagnosis algorithms is lacking due to data protection issues and partial observability. To combat this inherent lack of data, we introduce a high-fidelity, physics-informed co-simulation of a common aircraft main-fuel-pump system modelled in \textsc{MATLAB/Simulink Simscape Fluids}. We also describe its generated time-series data with health and fault mode annotations. To show feasibility of our benchmark, we apply an unsupervised Recurrent Variational Autoencoder (RNN-VAE) for anomaly detection and a SOM-VAE for operating mode discretization, trained to  separate healthy and faulty conditions.
\end{abstract}

\keywords{aircraft fuel system \and anomaly detection \and benchmark \and cyber-physical systems \and dataset \and machine learning \and predictive maintenance \and \textsc{Simscape} fluids \and time-series}

\section{Introduction}
Modern aircraft are filled with Cyber-Physical Systems (CPS) that process information and directly act on physical processes. In aerospace fuel systems, partial observability is common. Safety is guaranteed by certification, redundancies, scheduled maintenance and operational controls. However, many condition-monitoring sensors that are supplementary for operation are often not implemented as they impose additional cost on manufacturers \citep{Biondani2024}. Moreover, real fault data in the avionics domain is rare, restricting data-driven analysis \citep{Liu2025}. This increases testing time, thus failing to take advantage of predictive maintenance with trustworthy Artificial Intelligence (AI) \citep{Windmann2024}.

Because real operational datasets are effectively unavailable, researchers rely on high-fidelity benchmarks if they want to design, compare and validate anomaly detection or diagnosis models in the avionic domain. Therefore, we present a publicly available aircraft fuel system benchmark based on a high-fidelity, fault-injectable \textsc{Simscape} co-simulation. The simulator produces labelled multivariate time series, including noise, randomness and datasets with faults. In literature, even if simulations exist, they rarely have the type of labels that diagnosis algorithms require as input \citep{Li2023}, or other issues like few fault variety or dull system complexity \citep{ehrhardt2022benchmark}. 
To evaluate the proposed benchmark, algorithms for anomaly detection and system discretization are applied. 

The contributions of the paper are: (i) the novel design of a detailed aircraft fuel system in \textsc{MATLAB Simulink Simscape Fluids} with explicit fault injections for the main gear pump and related components. The model provides faults and allows to design controlled experiments; (ii) a labelled benchmark dataset containing time-series data
with health and fault mode annotations is published that is explicitly designed for an avionic domain where real operational data is typically unavailable. The datasets are publicly accessible to support reproducible evaluations of anomaly detection and diagnosis models at \url{https://github.com/Elix96J/Avionic_Main_Fuel_Pump_Simulation_and_Fault-Diagnosis_Benchmark}

As an exemplary use case, we evaluate the benchmark with a state-of-the-art unsupervised Recurrent Variational Autoencoder (RNN-VAE) for anomaly detection and discretize the proposed data to alter time-series into interpretable system descriptions \citep{overloper2024discretization}.
\section{Related Work}
Existing benchmarks in the context of CPS are highlighted by their strengths and limitations in supporting Consistency Based Diagnosis (CBD). This analysis sets the stage for introducing our proposed benchmark, which addresses gaps by offering a scalable, interpretable environment with diverse fault scenarios and explicit system model.

\subsection{Requirements for a Benchmark}
\cite{ehrhardt2022benchmark} provide the BeRfiPl benchmark for AI methods in anomaly diagnosis, reconfiguration and planning of Cyber-Physical Production Systems (CPPS). The benchmark consists of 16 simulated modular process-plant setups for different processes modelled in OpenModelica. On this basis, the authors propose a minimal set of five requirements for AI benchmarks for comparison:
\begin{itemize}
    \item Benchmarks should include both continuous and discrete signals \citep{Mercorelli2024},
    \item a variety of multiple types of faults \citep{balzereit2021Ensemble},
    \item high system complexity in multiple layers (e.g. interconnections, number of variables, fault scenarios) to test asymptotic performance of algorithms,
    \item non-trivial dependencies between inputs, outputs, parameters and other elements  \citep{research_agenda},
    \item repeating module types of a CPPS, ideally in loops, for the evaluation of planning paths algorithms to reuse functionalities \citep{research_agenda}.
\end{itemize}
\begin{table*}[!b]
\begin{center}
\caption{Comparison of conducted benchmarks}\label{tb:benchmarks}
\begin{tabular}{lcccccc}
    \toprule
    Benchmark & \emph{Model} & \emph{Hybrid signals} & \emph{Fault variety}  & \emph{Scalable} & \emph{System knowledge} & \emph{CBD-ready} \\
    \midrule
    BeRfiPl & Yes & Yes & No & Partially & Modules & No \\
    HAI-CPPS & Yes & Yes & Yes  & Yes & States \& residual modes & Yes \\
    TEP-Alarm  & Yes & Yes & Yes & Partially & Symbolic states & Partially \\
    Water Networks & Yes & No (continuous) & Yes & Yes &  Structural equations \& graph & Yes \\
    NoBOOM & No & Yes & Yes & Partially & Real data \& anomaly labels & No \\
    LiU-ICE & Yes & No (continuous) & No & Partially & Structural model \& DAE & No \\ \midrule
    This work & Yes & Yes & Yes & Yes & Health annotations & Yes \\
    \bottomrule
\end{tabular}
\end{center}
\end{table*}

Building on this foundation, \cite{Moddemann2025HAI-CPPS} extend BeRfiPl with the HAI-CPPS benchmark, which reuses the simulation but alters it for neuro-symbolic use cases. While \cite{ehrhardt2022benchmark} defines their fifth requirement as recurring module involvement, \cite{Moddemann2025HAI-CPPS} instead introduce a requirement for interpretable system topologies with explicit state information, providing symbolic representations. 
BeRfiPl offer limited fault cases, whereas HAI-CPPS extends the fault scenarios.

Table \ref{tb:benchmarks} summarizes the benchmark requirements from the literature and how the surveyed benchmarks, including our work, satisfy them. A \emph{model} indicates whether the benchmark provides an explicit representation or formalization of the CPS. This representation can be in the form of mathematical models, structural descriptions or formal models that describe system behavior. \emph{Hybrid signals} indicate the presence of both continuous and discrete variables. \emph{Fault variety} captures the diversity and complexity of the faults represented in the benchmark. A \emph{scalable} benchmark allows for the systematic variation of system complexity along multiple dimensions, including the number of variables and components, interconnection complexity between modules and fault scenario complexity. This is essential for studying the asymptotic performance limits of AI methods. \emph{System knowledge} refers to interpretable system descriptions, including topology information, behavioral models and symbolic representations of system states. Lastly, a \emph{CBD-ready} benchmark provides data containing fault annotations of faulty components and labelled discretisation of operating modes.

\subsection{Comparable Benchmarks}
\cite{manca2020tennessee} published a dataset that uses the Simulink model of \cite{Bathelt2015} to simulate process and alarm data based on the Tennessee-Eastman-Process (TEP) benchmark of \cite{downs1993plant}. \cite{manca2020tennessee} develops an alarm design in five successive revisions. The resulting TEP-Alarm dataset provides a large number of observations for alarm prediction. 
The TEP is considered partially scalable as it is tied to a single configuration and a fixed control structure, whereas disturbance scenarios and alarm thresholds vary between tests.
This creates challenges for anomaly detection as additional preprocessing steps are required to discretize alarms or deciding whether to predict alarm sequences or the underlying injected disturbances. Additionally, faults are solely scripted as minor disturbances, like step changes or valve closures and not degradation faults such as gear wear.

\cite{sztyber2022water} propose a water network benchmark for model-based fault diagnosis based on structural analysis. They find structurally overdetermined equation subsets to build residuals.

Extending this approach, \cite{sztyber:hal-04327789} follow-up work selects a real-world network to simulate the water distribution of a city, with additional historical information. The structural model is extended with more detailed component-level representations including explicit fault modes for a more realistic diagnosis benchmark on fluids. Although  structural models are published, immediately usable, logged and labelled time series are not delivered. Users must select sensors and leak locations before simulating separately if they want data, residuals and states.

\cite{wagnernoboom} introduce NoBOOM, a benchmark for anomaly detection in real-world chemical processes. Datasets of varying size and complexity are published, which are split into training and test sets, differing in  number of features, time steps and anomaly types, labelled by experts. However, detailed metadata is under non-disclosure and sensor IDs are anonymized. 
While NoBOOM provides valuable data, it lacks an executable plant model, which limits the ability to trace anomalies back to their root causes. Furthermore, researchers cannot simulate new operating scenarios or systematically inject component-level faults. The benchmark also exhibits a comparatively high anomaly rate that deviates from the classical anomaly detection setting of rare events.

\cite{Jung2025} created the LiU-ICE benchmark using a turbocharged gasoline engine test setup that mimics real-world combustion engine conditions, using eight sensors to track signals. The researchers also developed a model of the engine's air path using Differential Algebraic Equations (DAEs). LiU-ICE is applicable for residual-based diagnosis design for combustion engines but no explicit health annotations or residuals as the benchmark challenges participants to derive them. The benchmark covers only three sensor faults and a single leak fault and is not scalable as it relies on a fixed setup.

\section{A Novel High-Fidelity Simulation on Aircraft Fuel Systems}
\subsection{Simulation Setup}
Since Eaton is one of the major manufacturers of aircraft hydraulic pumps, we simulate Eaton main gear pumps. In practice, detailed geometric or measurement data is unavailable. To create a pump model under these constraints, the required pump parameters are reverse-engineered from the available catalogue data sheet and treated as a generic pump and parametrised by the properties provided \citep{eaton_main_engine_fuel_pump}. Specifically, five model types (704300, 708300, 708400, 708600, 714900) are simulated by a single pump, since they share similar characteristics. We implement fluid properties in \textsc{Simscape Fluids} via a custom Isothermal Liquid (IL) definition. This procedure can be applied directly to other manufacturers for which only catalogue data is available.
The simulation setup, see Figure ~\ref{fig:sim_overview}, is based on the configuration of \cite{Sciatti2022} and consists of the following components:
\begin{itemize}
    \item Tank/feeding system with $2\,\mathrm{bar}$ pre-pressurization, simulating the boost stage;
    \item Kerosene (Jet Fuel) with a density of $\rho = 784\,\mathrm{kg/m^3}$ and dynamic viscosity $\mu = 7.67 \times 10^{-4}\,\mathrm{Pa\cdot s}$, corresponding to a kinematic viscosity of $\nu = 9.8 \times 10^{-7}\,\mathrm{m^2/s}$ and an isothermal bulk modulus of $1.5\,\mathrm{GPa}$ at atmospheric pressure; 
    \item Main gear pump, modelled as a fixed-displacement, with discharge pressure $p_{\mathrm{out}} = 69\,\mathrm{bar} \,(1\,000\,\mathrm{psig})$ with an inlet pressure $p_{\mathrm{in}} = 2\,\mathrm{bar}\,(30\,\mathrm{psig})$ so that the pump increases the system pressure by 
    \begin{equation}
      \Delta p = p_{\mathrm{out}} - p_{\mathrm{in}} = 67\,\mathrm{bar}.
    \end{equation}
    while the rotational speed remains its nominal state with $n = 6\,000\,\mathrm{rev/min}\,(628.31\,\mathrm{rad/s})$ and a fuel flow rate of $Q = 58\,\mathrm{gpm} = 3.67 \times 10^{-3}\,\mathrm{m}^3/\mathrm{s}$ which translates into a pump displacement, given by:
    \begin{equation}
    D = \frac{Q}{\eta_{\mathrm{vol}} \cdot n} = 40.7\,\mathrm{cm}^3/\mathrm{rev}
    \end{equation}
    with an assumed $\eta_{\mathrm{vol}}=0.9$;
    \item Bypass valve with a pressure differential of $67\,\mathrm{bar}$, a pressure regulation range of $3.45\,\mathrm{bar}$ and a maximum opening area of $A_{\mathrm{Bypass,max}} \approx \pi \cdot 10\,\mathrm{mm} \cdot 5\,\mathrm{mm} \approx 157\,\mathrm{mm}^2$
    \citep{AMTS_Fokke};
    \item Fuel Metering Unit (FMU), modelled as a controllable orifice using the \texttt{Orifice} block, parameterised with a maximum geometric area of $A_{\mathrm{FMU,max}} = 100\,\mathrm{mm^2}$.  The metering area is controlled by a throttle, operated by the pilot to accelerate or decelerate the aircraft. The throttle signal $\theta(t)\in [0,1]$ represents a piecewise linear throttle variation:
    \begin{equation}
    \theta(t) =
    \left\{
    \begin{array}{ll}
    0.20, & 0 \le t \le 12, \\[4pt]
    0.20 + \frac{7}{30}(t - 12), & 12 < t \le 15, \\[4pt]
    0.90, & 15 < t \le 17, \\[4pt]
    0.90 - \frac{1}{5}(t - 17), & 17 < t \le 20, \\[4pt]
    0.30, & 20 < t \le T_{\mathrm{sim}};
    \end{array}
    \right.
    \end{equation}
    \item PRV opens at $200\,\mathrm{bar}$, connected from the pump discharge line to the tank and prevents the system from overpressurizing;
    \item Injector nozzles in front of the combustion engine with a total area of $25\,\mathrm{mm^2}$.
\end{itemize}

\begin{figure*}[!h]
  \centering
    \includegraphics[width=0.9\textwidth]{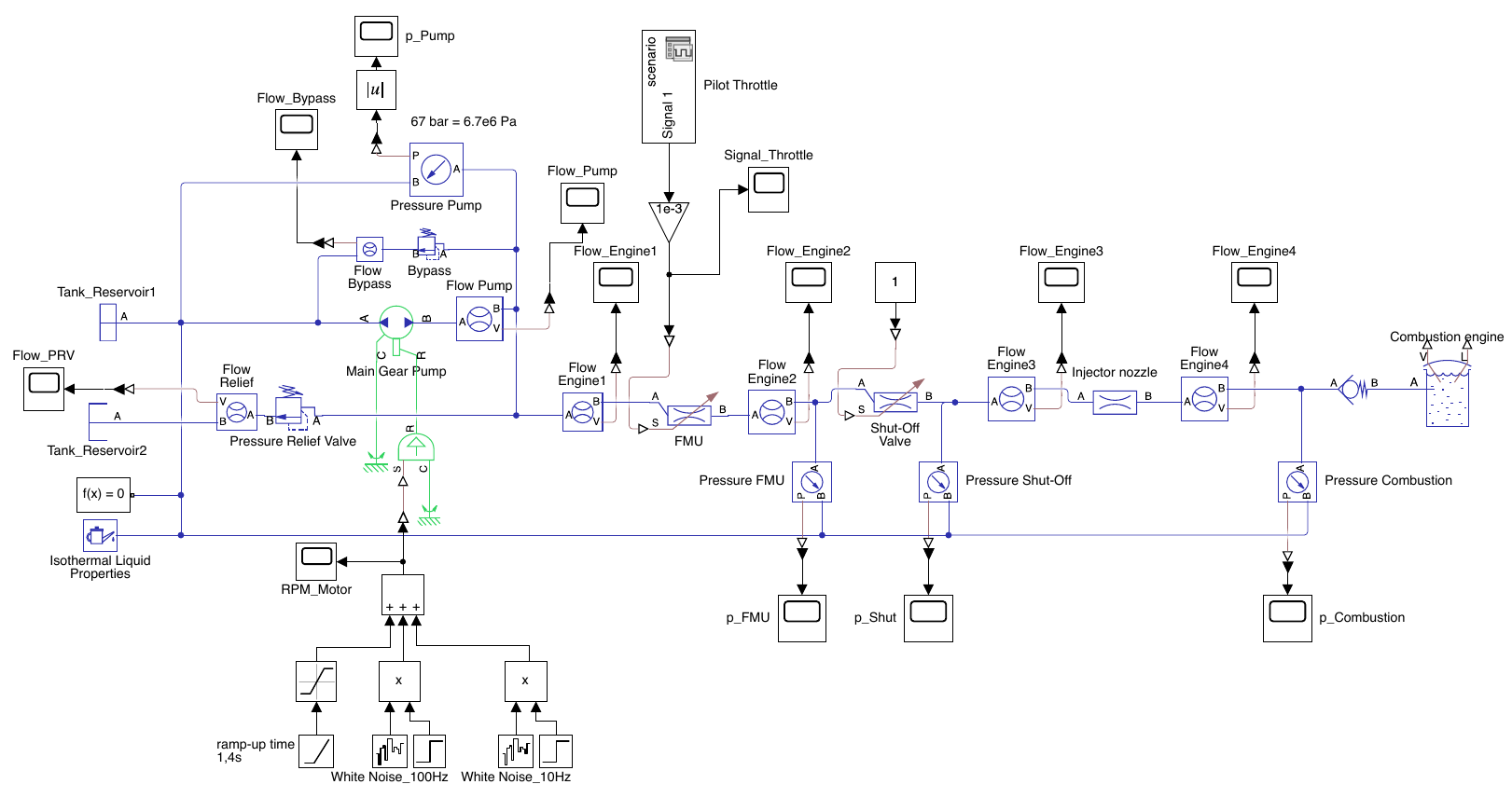}
  \caption{\textsc{Simulink/Simscape} model of the main fuel pump system}
  \label{fig:sim_overview}
\end{figure*}

\subsection{Fuel Flow Partitioning Methodology}

To dimension the fuel system, the flow has to be partitioned between the bypass branch and the path to the combustion engine. For estimating the distribution of fuel between the engine and bypass branches in the modelled fuel system, we scaled the flow based on the experimental configuration reported by \citep{Sciatti2022}. In their setup, the main fuel pump delivers a total flow rate of $Q_\mathrm{Pump,Sciatti} = 15\,\mathrm{L/min}$, with the bypass valve discharging $Q_\mathrm{Bypass,Sciatti} = 14\,\mathrm{L/min}$ at a piston position of $x_b = 1.1\,\mathrm{mm}$ and $Q_\mathrm{Bypass,Sciatti} = 2.8\,\mathrm{L/min}$ at $x_b = 0.1\,\mathrm{mm}$. Under these conditions the bypass carries between approximately $19\,\%$ and $93\,\%$ of the total pump flow, while the engine branch receives the remaining $7\,\%$ to $81\,\%$.

Assuming similar geometric values, these were scaled proportionally to the nominal pump flow rate of the present study, which is 
$ Q_\mathrm{Pump} \approx 2.4 \times 10^{-3}\,\mathrm{m^3/s}=144\,\mathrm{L/min}.$
Applying a scaling factor of
\begin{equation}
  k_{Q} = \frac{Q_\mathrm{Pump}}{Q_\mathrm{Pump,Sciatti}} = \frac{144}{15} \approx 9.6,
\end{equation}
the corresponding bypass volume flow rate lies in the range $Q_\mathrm{Bypass} \in [26.3, 131.6]\,\mathrm{L/min},$
where the lower bound corresponds to the nearly closed bypass position and the upper bound to the fully open bypass position. Using the continuity equation
\begin{equation}
  Q_\mathrm{Pump} = Q_\mathrm{Engine} + Q_\mathrm{Bypass},
\end{equation}
the engine volume flow rate lies in the range $ Q_\mathrm{Engine} \in \left[12.4,\,117.7\right]\,\mathrm{L/min}$.

Consistent with this scaling, the metering and restriction areas in the present model are chosen as scaled counterparts of the orifice geometries. The total injector nozzle area increases from $2.5\,\mathrm{mm^2}$ to $A_\mathrm{inj} = 25\,\mathrm{mm^2}$ and both the shut-off valve and the FMU metering valve are assigned a maximum effective area of $A_{\mathrm{SO,max}} = A_{\mathrm{FMU,max}} = 100\,\mathrm{mm^2}$. The bypass valve is dimensioned according to its spool geometry with piston diameter $D = 10\,\mathrm{mm}$ and maximum lift $x_{\max} = 5\,\mathrm{mm}$, leading to a maximum circumferential opening area
\begin{equation}
  A_{\max,\mathrm{geom}} \approx \pi \, D \, x_{\max} 
\end{equation}
\[
    \Rightarrow A_{\max,\mathrm{geom}} = \pi \cdot 10\,\mathrm{mm} \cdot 5\,\mathrm{mm} \approx 157\,\mathrm{mm^2},
\]
which is used as the maximum effective area of the bypass valve. With these properties, the model preserves the qualitative flow partitioning behaviour observed by \citep{Sciatti2022} while generating the higher nominal pump flow rate and pressure of an Eaton main gear pump.

\subsection{Simulating Faulty and Healthy Data}
For data generation, the hydraulic system is simulated in healthy and single-fault modes, resulting in labelled multivariate time series, see Figure ~\ref{fig:flow_pressure_plots}. Table \ref{table:faults} lists components $h$ and injected faults $f$. $h_0$ corresponds to the motor, $h_1$ to sensor, $h_2$ to the pump, $h_3$ to the PRV, $h_4$ to the FMU, $h_5$ to the bypass valve and $h_6$ to the fuel reservoir.
Components  $h_i \in \{\top,\bot\}$ are boolean with $\top$ denoting nominal behaviour and $\bot$ a faulty component \citep{diedrich2024learningSD}.
\begin{figure*}[!b]
  \centering
  \begin{subfigure}[t]{0.3\textwidth}
    \caption*{(i)}
    \includegraphics[width=\linewidth]{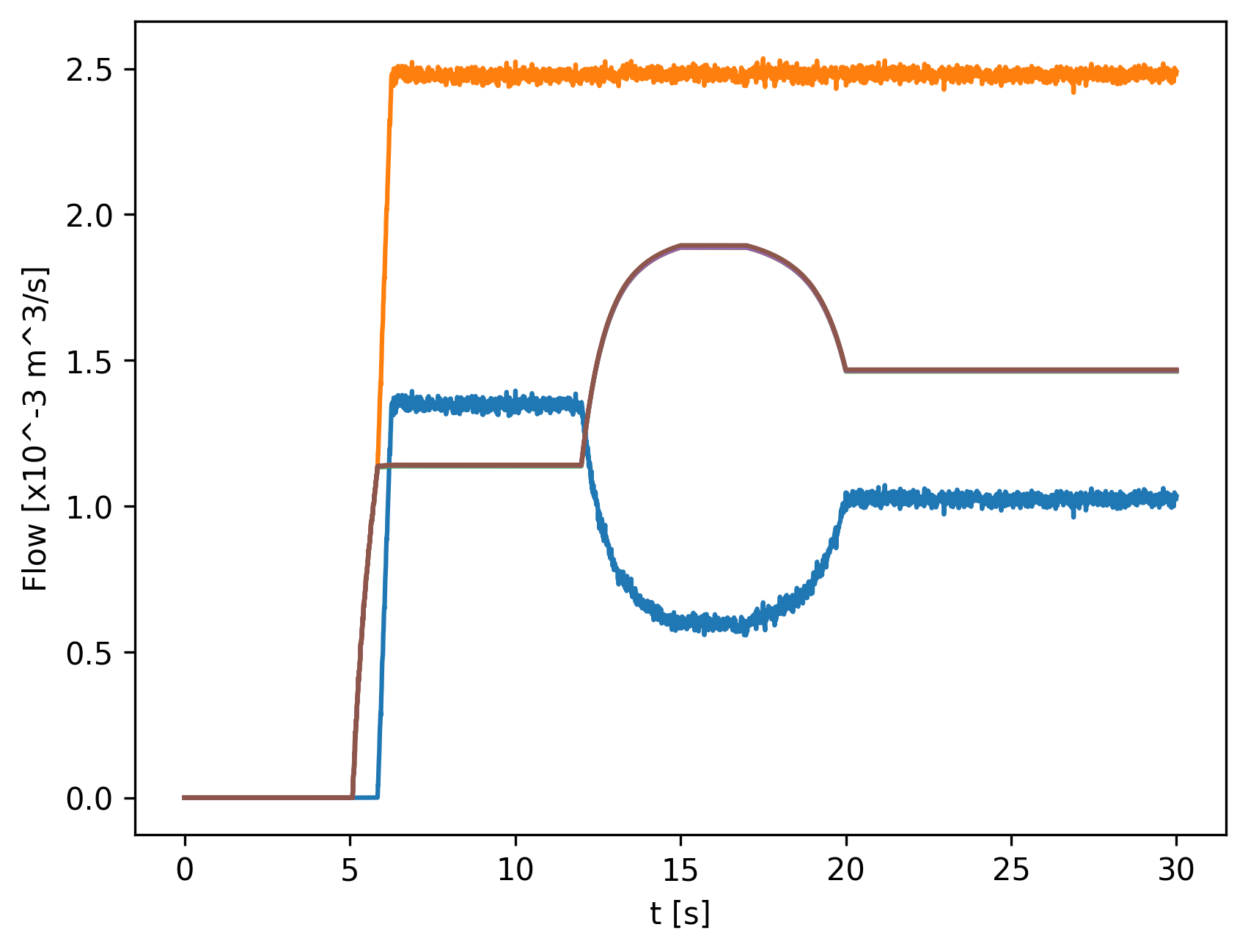}
    \includegraphics[width=\linewidth]{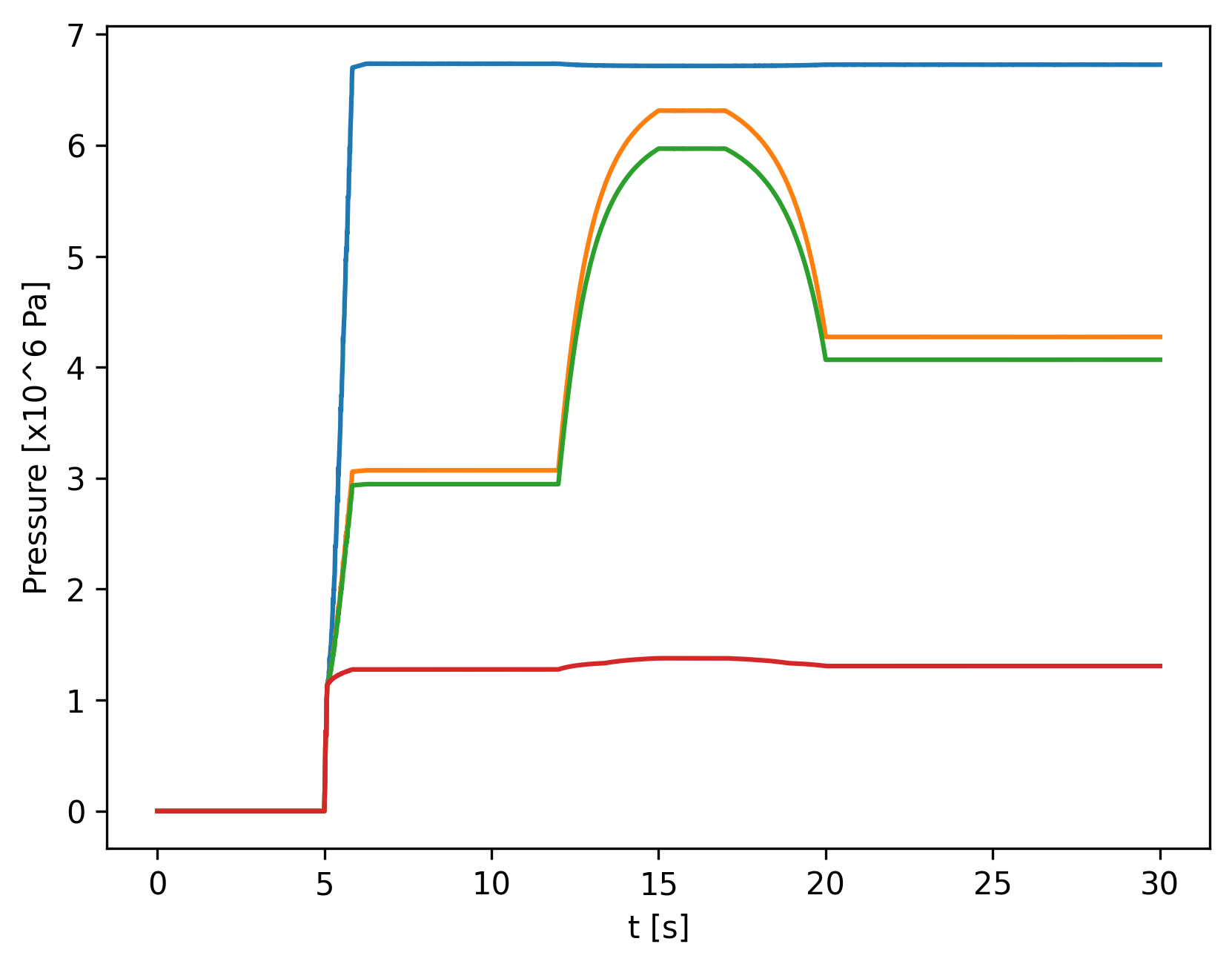}
  \end{subfigure}\hfill
  \begin{subfigure}[t]{0.3\textwidth}
    \caption*{(ii)}
    \includegraphics[width=\linewidth]{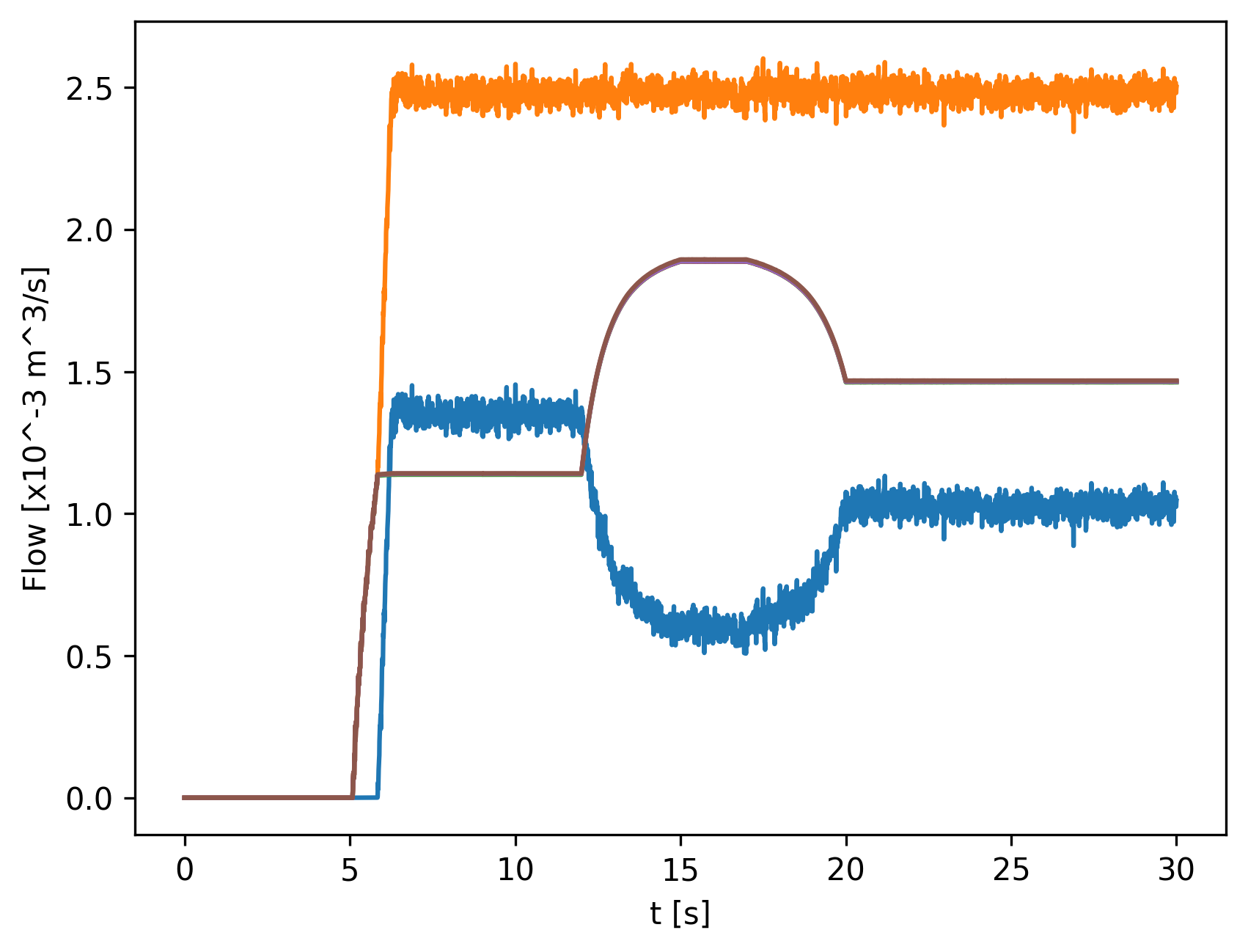}
    \includegraphics[width=\linewidth]{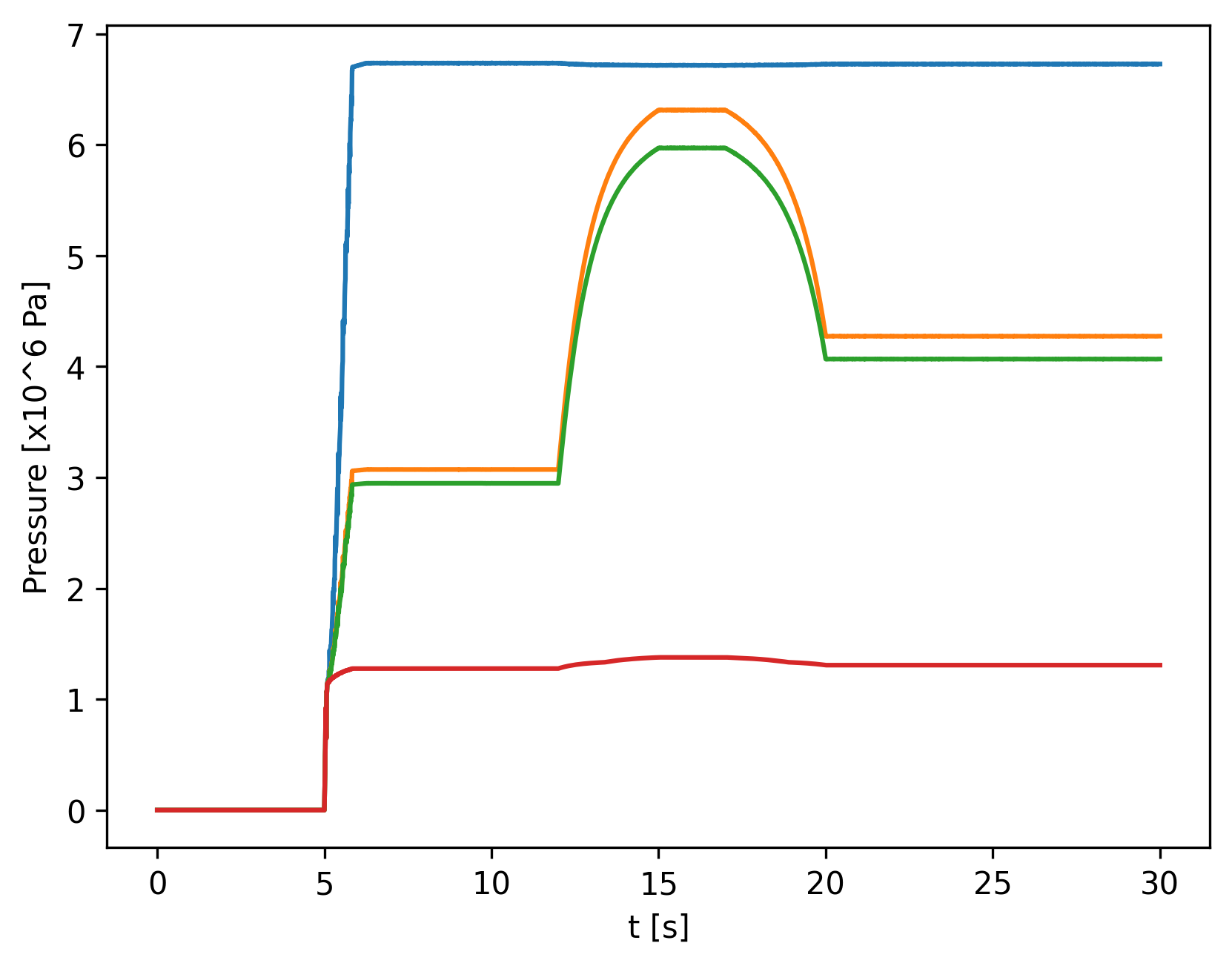}
  \end{subfigure}\hfill
  \begin{subfigure}[t]{0.3\textwidth}
    \caption*{(iii)}
    \includegraphics[width=\linewidth]{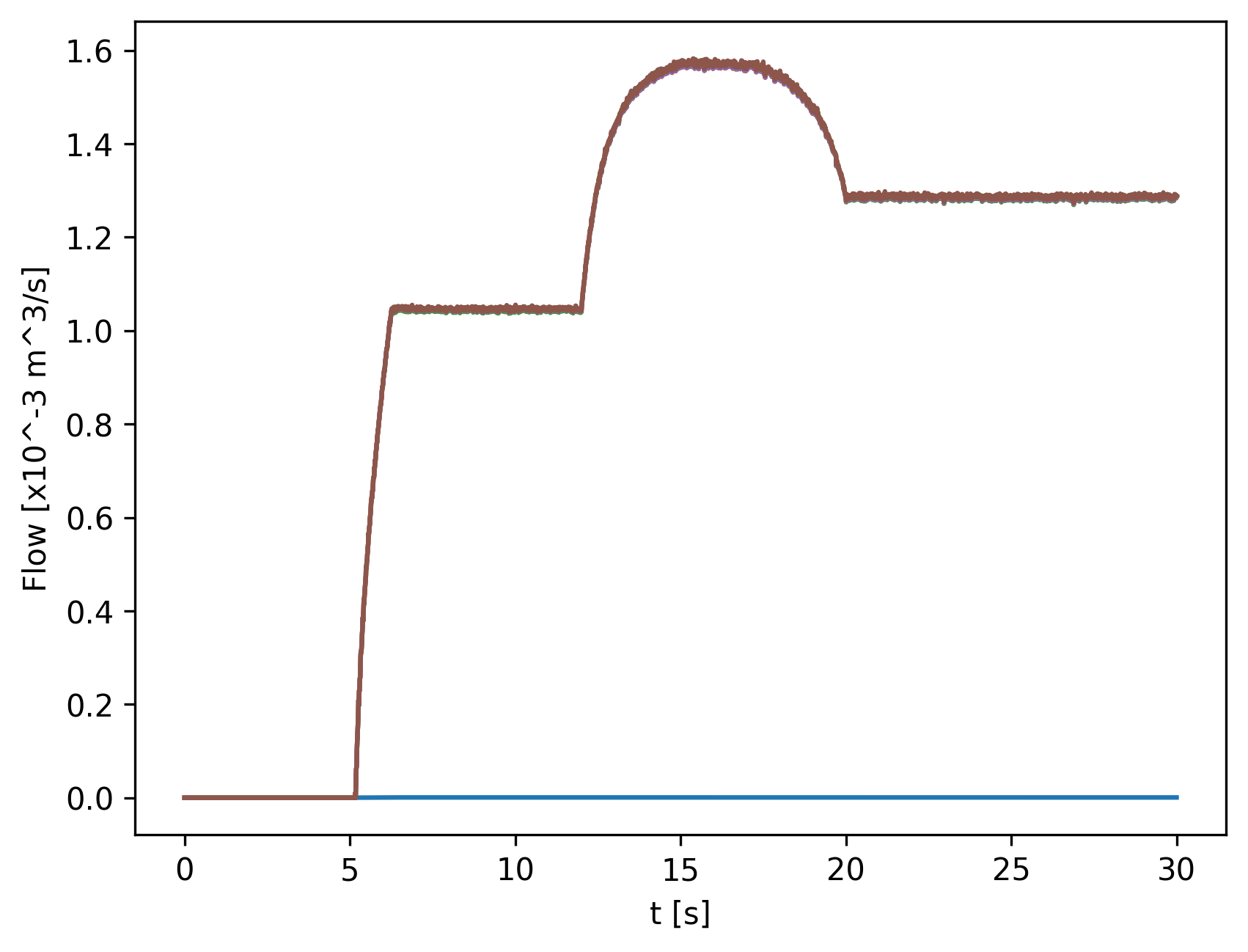}
    \includegraphics[width=\linewidth]{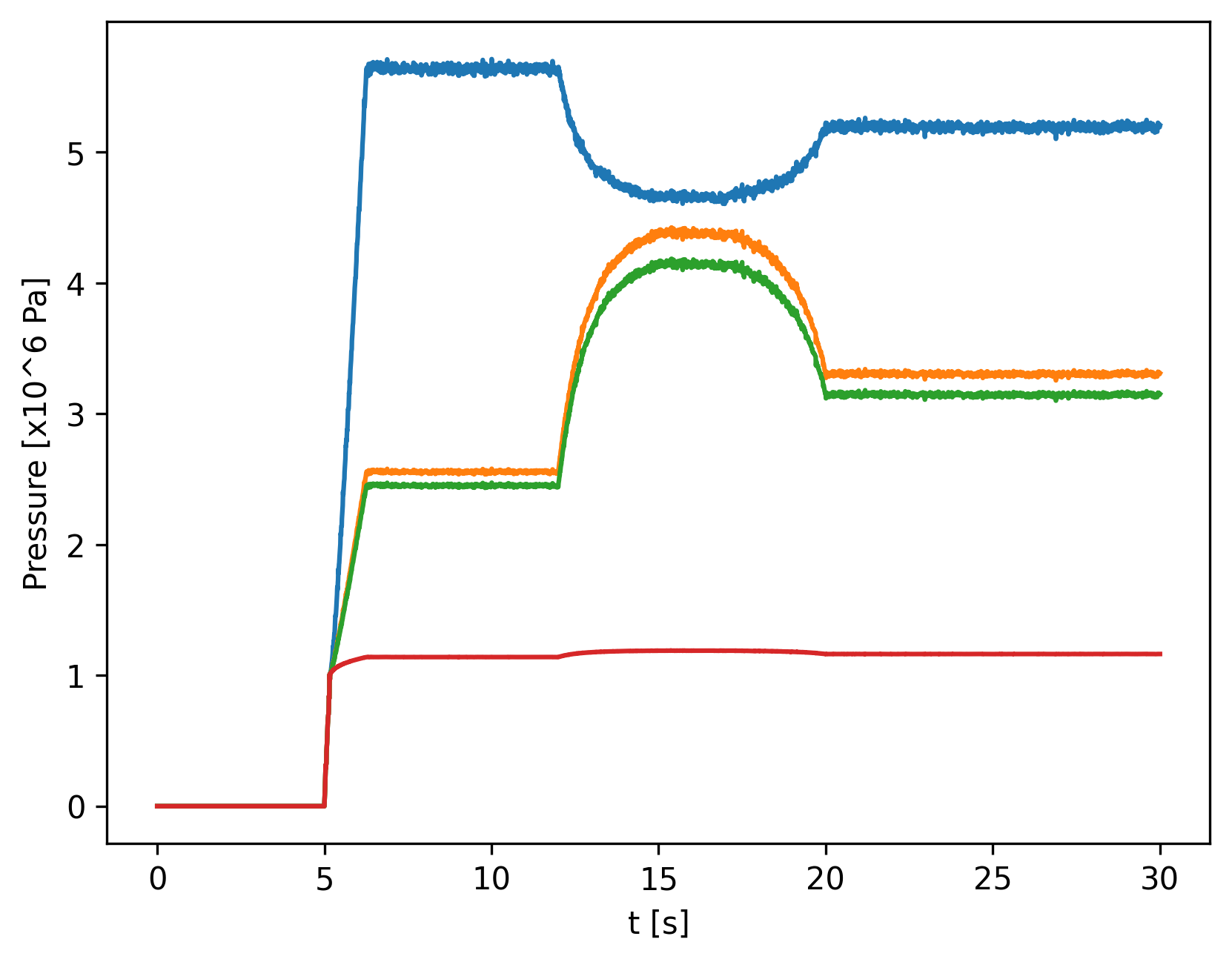}
  \end{subfigure}
  \caption{Time-series signals for the healthy\,(i) and two faulty runs\,(ii, iii). Top row displays flow rates with orange for $Q_\mathrm{Pump}$, wine red for $Q_\mathrm{Engine}$ and blue for $Q_\mathrm{Bypass}$. Bottom row displays pressure signals with blue for $p_\mathrm{Pump}$, orange for $p_\mathrm{FMU}$, green for $p_\mathrm{Shut}$ and red for $p_\mathrm{Combustion}$.}
\label{fig:flow_pressure_plots}
\end{figure*}

The global \textsc{Simulink} solver is set to \texttt{daessc}, a variable-step stiff DAE solver. Simulations are run in normal mode with the final simulation time specified via the \texttt{StopTime} parameter. All relevant signals are recorded using \textsc{Simulink} signal logging. Each run is saved as CSV file with a total number of $6\,001\,002$ rows, a simulation time of $30\,\mathrm{s}$, a sample rate $200\,\mathrm{Hz}$ and $1\,000$ simulation runs. 14 columns correspond to the number of recorded features plus an additional time column, while the healthy data file contains two additional columns for the Unique Identifiers (UD). In total, one file with only healthy data and nine files with faulty symptoms are generated.

\begin{table}[!h]
    \centering
    \caption{Faulty components $h_i$ for injected faults $f_i$}
    \label{table:faults}
    \vspace{2mm}
    \begin{tabular}{lcccccccc}
        \toprule
        $f$ fault 
        & $h_0$ 
        & $h_1$
        & $h_2$
        & $h_3$
        & $h_4$
        & $h_5$
        & $h_6$\\
        \midrule
        0 Healthy        & $\top$ & $\top$ & $\top$ & $\top$ & $\top$ & $\top$ & $\top$ \\
        1 White noise   & $\bot$ & $\top$& $\top$ & $\top$ & $\top$ & $\top$ & $\top$ \\
        2 Sensor drop $p_\mathrm{FMU}$
                           & $\top$ & $\bot$ & $\top$ & $\top$ & $\top$ & $\top$ & $\top$ \\
        3 Sensor drop $Q_\mathrm{Engine1}$ 
                           & $\top$ & $\bot$ & $\top$  & $\top$ & $\top$ & $\top$ & $\top$ \\
        4 Pump leakage
                           & $\top$ & $\top$ & $\bot$ & $\top$ & $\top$ & $\top$ & $\top$ \\
        5 Pump displacement change
                           & $\top$ & $\top$ & $\bot$ & $\top$ & $\top$ & $\top$ & $\top$ \\
        6 PRV leakage    & $\top$ & $\top$& $\top$ & $\bot$ &                                 $\top$ & $\top$ & $\top$ \\
        7 FMU orifice fault
                           & $\top$ & $\top$ & $\top$ & $\top$ & $\bot$ & $\top$ & $\top$ \\
        8 Bypass orifice fault
                           & $\top$ & $\top$ & $\top$ & $\top$ & $\top$ & $\bot$ & $\top$ \\
        9 High boost pressure 
                           & $\top$ & $\top$ & $\top$ & $\top$ & $\top$ & $\top$ & $\bot$ \\
        \bottomrule
    \end{tabular}
\end{table}

\section{Evaluation}

The proposed datasets are designed to meet the unique requirements of main-gear-pump testing. They are evaluated using task-specific ML methods for (i) anomaly detection and (ii) discretization to generate an interpretable, symbolic representation of the time-series. Each method is designed for its respective objective. These are robust model-building procedures that leverage explicit identifiers of model states and available health labels. Both methods confirm the immediate usability of the dataset.

\subsection{Anomaly Detection}

Detecting anomalous behavior in aviation component testing is complex, as it involves identifying infrequent or unpredictable events within operational data.
An unsupervised Variational Autoencoder (VAE) framework that uses the model likelihood of the observed data as a probabilistic residual for anomaly detection is applied. In particular, we adopt a Recurrent Variational Autoencoder (RNN-VAE) with a standard Gaussian prior distribution, $\mathcal{N}(0, 1)$, over the latent space, which facilitates more effective latent representation learning and consequently improves the model’s reconstruction capability for time-series.

Composite $F_\mathrm{1}$ score \citep{garg2021evaluation} is the harmonic mean of time-wise precision $Pr_t$ and event-wise recall $Rec_e$, thereby rewarding detectors that both avoid false alarms and reliably identify complete anomalous events, unlike standard $F_\mathrm{1}$ score. The model achieves high Composite $F_\mathrm{1}$ scores for most types of anomalies, with the exception of two cases, as listed in Table~\ref{tab:anomaly_detection_results}.

\begin{table}[ht]
\centering
\caption{Assessment of anomaly detection with Composite $F_\mathrm{1}$ (\checkmark \,fulfilled, \xmark\, not fulfilled)}
\vspace{2mm}
\begin{tabular}{lcc}
\toprule
$f$ fault & Detected & Comp. F1 Score\\
\midrule
1 White noise & \xmark & $0.0$ \\
2 Sensor drop $p_\mathrm{FMU}$ & \checkmark & $0.997$ \\
3 Sensor drop $Q_\mathrm{Engine1}$  & \checkmark & $0.994$ \\
4 Pump leakage & \checkmark & $0.999$ \\
5 Pump displacement change & \checkmark & $1.000$ \\
6 PRV leakage & \checkmark & $0.976$ \\
7 FMU orifice misposition & \checkmark & $0.994$ \\
8 Bypass orifice misposition & \xmark & $0.533$ \\
9 High boost pressure & \checkmark & $0.872$ \\
\bottomrule
\end{tabular}
\label{tab:anomaly_detection_results}
\end{table}

For the white-noise anomaly, the induced perturbation closely resembles the stochastic variability naturally present in nominal simulation data and is thereby difficult to detect. 
In the case of the bypass anomaly, the model has a high false-positive rate, which substantially reduces the corresponding anomaly detection performance.

\subsection{Discretization}
In the following, we demonstrate the utility of the newly developed dataset by deploying a SOM-VAE model \citep{overloper2024discretization} within a sequential input data setup for discretization and identify the thresholds in the latent space within which the algorithm reaches the trade-off between interpretability and its discretization performance.
To quantify the performance with respect to the discretization task, we adopt the metric Purity \(\pi\) by \cite{schutze2008introduction}, which serves as an appropriate measure for evaluating the alignment between predicted symbolic states and ground truth labels, formally defined as follows:
\begin{equation}
     \pi(C, S) = \frac{1}{n} \sum_{i=1}^{I} \max _{j}\left|s_{i} \cap c_{j}\right|
\end{equation}
where \(c_j \in C=\{c_1,…,c_J\}\) represent the true mode label, \(s_i \in S=\{s_1,…,s_J\}\) the set of symbolic modes generated by our model and \(n\) the total count of data samples.
Ground-truth labels are derived by one-hot encoding the pump’s operational phase (UD1) and its corresponding speed profile (UD2).
When combined, these two identifiers uniquely characterize the discrete states of the underlying automaton.
The discretization model is trained with a predefined number of latent categories i.e. SOM map units.
This number is systematically varied to assess its effect on the resulting evaluation metric, as summarized in Table \ref{tab:disc_results}.

\begin{table}[h]
\centering
\caption{Purity results for varying SOM map sizes and sequence lengths.}
\vspace{2mm}
\begin{tabular}{ccc}
\toprule
Latent SOM Map & Sequence Length & Result \\
\midrule
$[3,3]$   & 20  & 0.7988 \\
$[3,3]$   & 50  & 0.7611 \\
$[3,3]$   & 100 & 0.7352 \\
$[6,6]$   & 20  & 0.8540 \\
$[12,12]$ & 20  & 0.8846 \\
\bottomrule
\end{tabular}
\label{tab:disc_results}
\end{table}

The results demonstrate that the discretization mechanism maintains strong performance and benefits from increased representational capacity as the size of the SOM map grows.
This is to be expected, as larger maps enable the model to encode a greater amount of information and partition the latent space with higher granularity.
Conversely, performance decreases as the sequence length increases, indicating that longer temporal contexts introduce additional complexity that challenges the discretization.

\section{Conclusion}
Our datasets are constrained in realism and generalizability as they are synthetic, generated from a Simscape co-simulation. Only single-fault cases are covered and the throttle signal is a fixed, piecewise-linear signal. 
Further simplifications include the assumption of an isothermal fluid, disregarding temperature effects. Cavitation is not represented and no sensor latency is modelled as logged signals are interpreted to be ideal.
Nonetheless, the model and data are public, allowing to add multi-fault cases and adjust inputs, properties as well as components. Applying the RNN-VAE for anomaly detection and the SOM-VAE for discretization results in strong performance, which confirms that the datasets support learning and are appropriate for evaluating a broader class of models. The RNN-VAE detects most faults well, but completely misses the white-noise anomaly as it mimics gear wear. High false positives are shown on the bypass-orifice misposition, indicating signatures similar to the healthy sequence.

Further research involves the enhancement of the simulation model to include more details on the interaction between mechanical and electronic components in the hydraulic chain, implementing different pilot throttle signals, adding fault superpositions and implementing different ML prediction models for time-series to obtain better metrics and evaluate the benchmark. Additionally, validating the simulation against real pump measurements would further strengthen the credibility of the model.
\newpage
\bibliographystyle{unsrtnat}
\bibliography{references}

@article{downs1993plant,
	title={A plant-wide industrial process control problem},
	author={Downs, James J and Vogel, Ernest F},
	journal={Computers \& chemical engineering},
	volume={17},
	number={3},
	pages={245--255},
	year={1993},
	publisher={Elsevier}
}

@article{Li2023,
  title = {Intelligent Fault Diagnosis of an Aircraft Fuel System Using Machine Learning—A Literature Review},
  volume = {11},
  number = {4},
  journal = {Machines},
  publisher = {MDPI AG},
  author = {Li,  Jiajin and King,  Steve and Jennions,  Ian},
  year = {2023},
  pages = {481}
}

@inproceedings{Biondani2024,
  title = {Fault Injection for Synthetic Data Generation in Aircraft: A Simulation-Based Approach},
  booktitle = {2024 IEEE 22nd International Conference on Industrial Informatics (INDIN)},
  publisher = {IEEE},
  author = {Biondani,  Francesco and Dall’Ora,  Nicola and Tosoni,  Francesco and Fraccaroli,  Enrico and Migliore,  Domenico Fabio and Acerra,  Francesco and Fummi,  Franco},
  year = {2024},
}

@INPROCEEDINGS{research_agenda,
  author={Köcher, Aljosha and Heesch, René and Widulle, Niklas and Nordhausen, Anna and Putzke, Julian and Windmann, Alexander and Niggemann, Oliver},
  booktitle={2022 IEEE 5th International Conference on Industrial Cyber-Physical Systems (ICPS)}, 
  title={A Research Agenda for {AI} Planning in the Field of Flexible Production Systems}, 
  year={2022},
  volume={},
  number={},
  pages={1-8}
}

@InProceedings{balzereit2021Ensemble,
  author    = {Balzereit, Kaja and Diedrich, Alexander and Ginster, Jonas and Windmann, Stefan and Niggemann, Oliver},
  booktitle = {19th IEEE International Conference on Industrial Informatics},
  title     = {An Ensemble of Benchmarks for the Evaluation of {AI} Methods for Fault Handling in {CPPS}},
  year      = {2021},
  month     = {11}
}

@inproceedings{Windmann2024,
  title = {Artificial {I}ntelligence in {I}ndustry 4.0: A Review of Integration Challenges for Industrial Systems},
  booktitle = {2024 IEEE 22nd International Conference on Industrial Informatics (INDIN)},
  publisher = {IEEE},
  author = {Windmann,  Alexander and Wittenberg,  Philipp and Schieseck,  Marvin and Niggemann,  Oliver},
  year = {2024},
  pages = {1–8}
}

@article{Sciatti2022,
  title = {Detailed simulations of an aircraft fuel system by means of {S}imulink},
  volume = {2385},
  journal = {Journal of Physics: Conference Series},
  publisher = {IOP Publishing},
  author = {Sciatti,  Francesco and Tamburrano,  Paolo and De Palma,  Pietro and Distaso,  Elia and Amirante,  Riccardo},
  year = {2022},
}

@inproceedings{manca2020tennessee,
    author    = {Manca, Gianluca},
    title     = {{Tennessee-Eastman-Process} Alarm Management Dataset},
    year      = {2020}, 
    publisher = {IEEE DataPort},
}

@inproceedings{wagnernoboom,
    title={NoBOOM: Chemical Process Datasets for Industrial Anomaly Detection},
    author={Wagner, Dennis and Hartung, Fabian and Arweiler, Justus and Muraleedharan, Aparna and Jungjohann, Indra and Nair, Arjun and Reithermann, Steffen and Schulz, Ralf and Bortz, Michael and Neider, Daniel and others},
    booktitle={The Thirty-ninth Annual Conference on Neural Information Processing Systems Datasets and Benchmarks Track},
    year      = {2020},
}

@inproceedings{Moddemann2025HAI-CPPS,
  title = {The {HAI-CPPS} Benchmark: Evaluating {AI} Capabilities across Hybrid Data Spaces},
  booktitle = {2025 IEEE 30th International Conference on Emerging Technologies and Factory Automation (ETFA)},
  publisher = {IEEE},
  author = {Moddemann,  Lukas and Ehrhardt,  Jonas and Diedrich,  Alexander and Niggemann,  Oliver},
  year = {2025},
  pages = {1–8}
}

@inproceedings{ehrhardt2022benchmark,
  title={An {AI} benchmark for Diagnosis, Reconfiguration \& Planning},
  author={Ehrhardt, Jonas and Ramonat, Malte and Heesch, Rene and Balzereit, Kaja and Diedrich, Alexander and Niggemann, Oliver},
  booktitle={2022 IEEE 27th International Conference on Emerging Technologies and Factory Automation (ETFA)},
  year={2022},
  organization={IEEE}
}

@manual{AMTS_Fokke,
  title        = {Fokker 50/60 Technical Training: 73. Engine Fuel and Control — Engine Fuel System: Description and Operation},
  author = {Aircraft Maintenance \& Training School (AM\&TS) \& Fly Fokker},
  organization = {Aircraft Maintenance \& Training School (AM\&TS) \& Fly Fokker},
  year         = {2010},
  type         = {Training manual},
  note         = {ATA 73}
}

@inproceedings{sztyber2022water,
  title={Water network benchmarks for structural analysis algorithms in fault diagnosis},
  author={Sztyber, Anna and Chanthery, Elodie and Trav{\'e}-Massuy{\`e}s, Louise and P{\'e}rez-Zu{\~n}iga, Carlos Gustavo},
  booktitle={33rd International Workshop on Principle of Diagnosis--DX 2022},
  year={2022}
}

@inproceedings{sztyber:hal-04327789,
  TITLE = {Benchmark for Fault Diagnosis of Water Distribution Network},
  AUTHOR = {Sztyber, Anna and Chanthery, Elodie and Trav{\'e}-Massuy{\`e}s, Louise},
  BOOKTITLE = {{The 34th International Workshop on Principles of Diagnosis (DX'23)}},
  ADDRESS = {Loma Mar, United States},
  ORGANIZATION = {{PARC and the Silicon Austria Labs (SAL).}},
  YEAR = {2023}
}

@misc{eaton_main_engine_fuel_pump,
  author = {{Eaton Corporation}},
  title = {Main Engine Fuel Pump},
  year = {2025},
  note = {Accessed: 2025-08-10},
  url = {https://www.eaton.com/content/dam/eaton/products/pumps/aerospace-fuel-pumps/documents/eaton-main-gear-fuel-pump-brochure-tf600-52_en-us.pdf}
}

@article{Jung2025,
  title = {A fault diagnosis benchmark of technical systems with incomplete data — six solutions},
  volume = {164},
  ISSN = {0967-0661},
  journal = {Control Engineering Practice},
  publisher = {Elsevier BV},
  author = {Jung,  Daniel and Frisk,  Erik and Krysander,  Mattias and Sztyber-Betley,  Anna and Corrini,  Francesco and Arici,  Andrea and Anselmi,  Nicolas and Mazzoleni,  Mirko and Xu,  Jiamin and Mo,  Siwen and Xu,  Zixuan and Yang,  Chongpan and Du,  Zhile and Safaeipour,  Hossein and Forouzanfar,  Mehdi and Mirahi,  Vahid and Pinnarelli,  Anna and Puig,  Vicen\c{c} and Deng,  Qiao and Liu,  Yufei and Liu,  Jiakun and Ke,  Haobin and Zhu,  Wanting and Merkelbach,  Silke and Ahang,  Maryam and Najjaran,  Homayoun},
  year = {2025},
  pages = {106427}
}

@article{Mercorelli2024,
  title = {Recent Advances in Intelligent Algorithms for Fault Detection and Diagnosis},
  volume = {24},
  ISSN = {1424-8220},
  number = {8},
  journal = {Sensors},
  publisher = {MDPI AG},
  author = {Mercorelli,  Paolo},
  year = {2024},
  month = apr,
  pages = {2656}
}

@article{Liu2025,
  title = {Aviation Fuel Pump Fault Diagnosis Based on Conditional Variational Self-Encoder Adaptive Synthetic Less Data Enhancement},
  volume = {13},
  number = {14},
  journal = {Mathematics},
  publisher = {MDPI AG},
  author = {Liu,  Tiejun and Zhang,  Yaoping and Yin,  Xiaojing and He,  Weidong},
  year = {2025},
  pages = {2218}
}

@article{Bathelt2015,
  title = {Revision of the {T}ennessee {E}astman {P}rocess Model},
  volume = {48},
  ISSN = {2405-8963},
  number = {8},
  journal = {IFAC-PapersOnLine},
  publisher = {Elsevier BV},
  author = {Bathelt,  Andreas and Ricker,  N. Lawrence and Jelali,  Mohieddine},
  year = {2015},
  pages = {309–314}
}

@inproceedings{diedrich2024learningSD,
  title={Learning System Descriptions for Cyber-Physical Systems},
  author={Diedrich, Alexander and Moddemann, Lukas and Niggemann, Oliver},
  booktitle={Proceedings of 12th IFAC Symposium on Fault Detection, Supervision and Safety for Technical Processes},
  year={2024}
}

@inproceedings{overloper2024discretization,
  title={Discretization of {CPS} Time Series with Neural Networks},
  author={Overl{\"o}per, Phillip Johann and Moddemann, Lukas and Hranisavljevic, Nemanja and Windmann, Alexander and Niggemann, Oliver},
  booktitle={2024 IEEE 29th International Conference on Emerging Technologies and Factory Automation (ETFA)},
  pages={1--8},
  year={2024},
  organization={IEEE}
}

@book{schutze2008introduction,
  title={{Introduction to Information Retrieval}},
  author={Manning, Christopher D},
  year={2008},
  publisher={Syngress Publishing}
}

@article{garg2021evaluation,
  title={{An Evaluation of Anomaly Detection and Diagnosis in Multivariate Time Series}},
  author={Garg, Astha and Zhang, Wenyu and Samaran, Jules and Savitha, Ramasamy and Foo, Chuan-Sheng},
  journal={IEEE Transactions on Neural Networks and Learning Systems},
  volume={33},
  number={6},
  pages={2508--2517},
  year={2021},
  publisher={IEEE}
}
\end{document}